\pdfoutput=1
\documentclass[11pt]{article}

\usepackage[]{ACL2023}

\usepackage{times}
\usepackage{latexsym}
\usepackage{booktabs}

\usepackage[T1]{fontenc}
\usepackage[utf8]{inputenc}

\usepackage{microtype}

\usepackage{inconsolata}
\usepackage[utf8]{inputenc} % allow utf-8 input
\usepackage[T1]{fontenc}    % use 8-bit T1 fonts
\usepackage{natbib}
\usepackage{multicol}
\usepackage{multirow}
\usepackage{hyperref}       % hyperlinks
\usepackage{url}            % simple URL typesetting
\usepackage{booktabs}       % professional-quality tables
\usepackage{amsfonts}       % blackboard math symbols
\usepackage{nicefrac}       % compact symbols for 1/2, etc.
\usepackage{microtype}      % microtypography
\usepackage{xcolor}         % colors
\usepackage{changepage}
\usepackage{graphicx}

\definecolor{hotpink}{RGB}{255, 83, 115}
\definecolor{teal}{RGB}{90, 200, 250}
\definecolor{lightgreen}{RGB}{33, 222, 128}
\definecolor{lightblue}{RGB}{72, 123, 232}
\definecolor{darkgreen}{RGB}{56, 117, 79}

\title{Adversarial Robustness of Prompt-based Few-Shot Learning for Natural Language Understanding}

\author{Venkata Prabhakara Sarath Nookala$^*$\\
  Georgia Institute of Technology \\
  \texttt{vnookala3@gatech.edu} \\\And
  Gaurav Verma$^*$ \\
  Georgia Institute of Technology \\
  \texttt{gverma@gatech.edu} \\
  \AND
  Subhabrata Mukherjee \\
  Microsoft Research \\
  \texttt{subhabrata.mukherjee@microsoft.com} \\\And
  Srijan Kumar \\
  Georgia Institute of Technology \\
  \texttt{srijan@gatech.edu}  \\
  }

\begin{document}
\maketitle
\begingroup\def\thefootnote{*}\footnotetext{Equal contribution.}\endgroup
\begin{abstract}
  State-of-the-art few-shot learning (FSL) methods leverage prompt-based fine-tuning to obtain remarkable results for natural language understanding (NLU) tasks. While much of the prior FSL methods focus on improving downstream task performance, there is a limited understanding of the adversarial robustness of such methods. In this work, we conduct an extensive study of several state-of-the-art FSL methods to assess their robustness to adversarial perturbations. To better understand the impact of various factors towards robustness (or the lack of it), we evaluate prompt-based FSL methods against fully fine-tuned models for aspects such as the use of unlabeled data, multiple prompts, number of few-shot examples, model size and type. Our results on six GLUE tasks indicate that compared to fully fine-tuned models, vanilla FSL methods lead to a notable relative drop in task performance (i.e., are less robust) in the face of adversarial perturbations. However, using \textit{(i)} unlabeled data for prompt-based FSL and \textit{(ii)} multiple prompts flip the trend.
  We further demonstrate that increasing the number of few-shot examples and model size lead to increased adversarial robustness of vanilla FSL methods. Broadly, our work sheds light on the adversarial robustness evaluation of prompt-based FSL methods for NLU tasks. 
\end{abstract}

\section{Introduction}
Few-shot learning (FSL) capabilities of large language models have led to a remarkable performance on several natural language understanding (NLU) tasks, often with as little as $16$ examples per class~\cite{mukherjee2021clues, lester2021power, li2021prefix, wang2021list}. Prompt-based few-shot learning is one such approach where NLU tasks are reformulated as prompts, which are then completed using large language models~\cite{gao2020making, schick2020s, tam2021improving, liu2021gpt}. By effectively bridging the gap between the pre-training objective of large language models and the fine-tuning objective, prompt-based learning has provided impressive results. Several recent studies have investigated conditioning large language models to solve downstream tasks by prompting them with a few examples. 

\begin{figure}
    \centering
    \includegraphics[width=\linewidth]{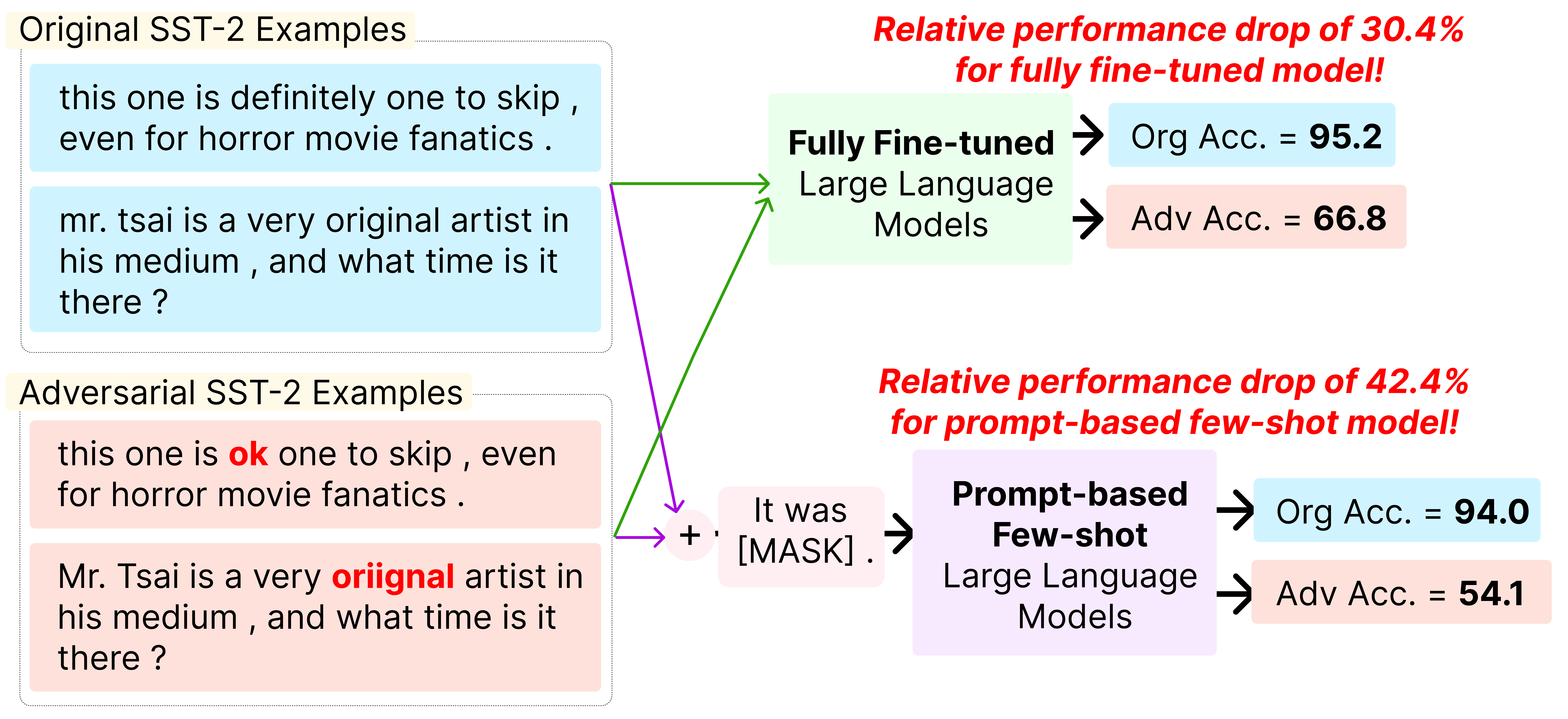}
    \caption{\textbf{Overview of our study.} We compare the relative gap between the in-domain and adversarial performance of different state-of-the-art prompt-based few-shot learning methods with that of the models trained with fully supervised learning.}
    \label{fig:overview}
\end{figure}

While much of the prior FSL works~\cite{gao2020making, liu2021gpt, tam2021improving, lester2021power} focus on improving downstream task performance, it is also critical to evaluate language technologies for adversarial robustness as that can highlight the security and safety risks of integrating them into user-sensitive downstream applications.
The robustness and generalization capabilities of prompt-based few-shot learning models have been the focus of some recent studies. For instance, ~\citeauthor{razeghi2022impact} (\citeyear{razeghi2022impact}) found that prompting language models is not robust to pre-training term frequencies in the context of arithmetic tasks. In a similar vein, a recent study found that prompt-based FSL is susceptible to learning superficial cues that hinder the generalizability of such methods~\cite{kavumba2022prompt}.
On the other hand, encouragingly, ~\citeauthor{liu2022sample} (\citeyear{liu2022sample}) and ~\citeauthor{awadalla2022exploring} (\citeyear{awadalla2022exploring}) found that prompt-based FSL leads to more robust models in the face of out-of-distribution samples. We add to the existing body of work by specifically studying the robustness of prompting to adversarial samples, which is different from studying the robustness against natural out-of-distribution samples---unlike natural distribution shifts, adversarial samples are carefully designed to exploit the vulnerabilities of language technologies and can pose serious safety concerns in real-world applications~\cite{madry2017towards, huang2017adversarial}.

In this work, we conduct the first study that empirically evaluates the adversarial robustness of prompt-based FSL methods for NLU and compares it against the robustness of fully supervised models. 
We study several state-of-the-art prompt-based FSL methods and evaluate their adversarial robustness on $6$ different tasks included in the GLUE benchmark. For each of the tasks, we use the adversarial evaluation set (AdvGLUE) curated by ~\citeauthor{wang2021adversarial} (\citeyear{wang2021adversarial}) to quantify the adversarial robustness of different FSL methods. AdvGLUE is a rich adversarial benchmark that comprises human-validated adversarially perturbed examples that include automated word- and sentence-level perturbations as well as human-crafted examples. We select prompt-based learning approaches that include the following modeling variations: \textit{(i)} no use of unlabeled data, \textit{(ii)} use of unlabeled data, and \textit{(ii)} use of multiple prompts for ensembling. Together, these modeling variations cover the different categories of prompt-based FSL methods identified in the FewNLU benchmark~\cite{zheng2021fewnlu}. Finally, we compare the models trained using prompting techniques with models trained on fully labeled data using conventional fine-tuning in terms of the gap in the performance between the adversarial and the in-domain evaluation sets. 

\noindent We summarize our findings below:\\
\textbf{1}. Vanilla prompt-based fine-tuning (LM-BFF~\cite{gao2020making}) demonstrates a worse relative drop in adversarial performance with respect to in-domain performance than full fine-tuning, and even classic fine-tuning with few examples.\\
\textbf{2}. However, using unlabeled data (iPET~\cite{schick2020s}) during fine-tuning flips the trend, causing prompting to reduce the drop in adversarial performance with respect to in-domain performance than full fine-tuning.\\
\textbf{3}. Similarly, using multiple prompts to fine-tune multiple models (PET~\cite{schick2020s}) and ensembling the resultant predictions cause prompting to demonstrate a better relative drop in adversarial performance with respect to in-domain performance than full fine-tuning.\\
\textbf{4}. Using several ablations, we demonstrate that increasing the number of few-shot examples and the encoder size reduces the relative drop in adversarial performance with respect to in-domain performance. We also find that RoBERTa~\cite{liu2019roberta} encoders are more adversarially robust than ALBERT~\cite{lan2019albert} and BERT~\cite{devlin2018bert} encoders of comparable size.\footnote{Code for our experiments: \url{https://github.com/claws-lab/few-shot-adversarial-robustness}}

We discuss the implications of these findings and contextualize them with respect to prior studies on other aspects of the robustness of prompt-based few-shot learning. 

\section{Related Work}
\noindent\textbf{Few-shot Learning for NLU}:
Few-shot learning aims to train models to perform well on a wide range of natural language understanding tasks with a small amount of task-specific training data~\cite{zheng2021fewnlu, mukherjee2021clues}. Recent studies have explored a wide range of techniques for few-shot learning, like meta-learning on auxiliary tasks~\cite{dou2019investigating, nooralahzadeh2020zero}, semi-supervised learning with unlabeled data~\cite{xie2020unsupervised, mukherjee2020uncertainty}, and intermediate learning with related tasks~\cite{yin2020universal, zhao2021calibrate, phang2018sentence}. A popular and influential branch of few-shot learning approaches involves fine-tuning large language models using \textit{prompting}~\cite{schick2021exploiting}. In such approaches, a handful of training examples are transformed using \textit{templates} and \textit{verbalizers}, and the language models are trained to predict the masked verbalizers under various settings.\footnote{For instance, the sentiment classification could involve the following transformation using a template: ``\textit{I loved the movie!}'' $\rightarrow$ ``\textit{I loved the movie! It was \texttt{[MASK]}}'', with the language models being trained to predict verbalizers ``great'' or ``terrible'' for positive and negative sentiment labels, respectively.} By framing the downstream tasks as a \texttt{MASK} prediction task, prompt-based learning overcomes the requirement of training task-specific classification heads, matching the fine-tuning objective with the pre-training objective. FewNLU~\cite{zheng2021fewnlu}, a benchmark designed to evaluate the performance of prompt-based few-shot learning capabilities systematically, categorizes these settings to fall in one or more of the following categories: \textit{(i)} not using any unlabeled data, \textit{(ii)} using unlabeled data, and \textit{(iii)} using an ensemble of models trained using different prompts.  Overall, evaluation of multiple prompt-based few-shot learning approaches has demonstrated that they solve NLU tasks to a remarkable extent with as little as $16$ labeled examples per class when compared against fine-tuned models that are trained with thousands of labeled examples. Such data-efficient learning capabilities are critical for building language technologies where it is challenging to collect large-scale labeled datasets. However, these approaches must demonstrate adversarial robustness to ensure safe outcomes in real-world applications where untrusted sources could supply the inputs. To this end, in this work, we systematically study the adversarial robustness of prompt-based few-shot learning approaches while considering the benefits of various settings identified in the FewNLU benchmark (i.e., the role of unlabeled data and ensembling). 

\vspace{0.05in}
\noindent\textbf{Robustness of  Few-shot Learning}:
Prior work has investigated the robustness of various few-shot learning of computer vision~\cite{goldblum2020adversarially} and natural language processing models~\cite{liu2022sample, awadalla2022exploring}, with some works also developing new robust learning approaches~\cite{jiang2019smart, wortsman2022robust}. Such robustness assessments are distinguished into two categories: \textit{(a)} robustness to natural and unintentional perturbations, and \textit{(b)} robustness to adversarial perturbations. Our work focuses explicitly on the adversarial robustness of prompt-based few-shot learning for natural language understanding.    

The most related works to ours are the studies by ~\citeauthor{liu2022sample} (\citeyear{liu2022sample}) and ~\citeauthor{awadalla2022exploring} (\citeyear{awadalla2022exploring}). Both studies consider the robustness of a wide range of data-efficient approaches to out-of-distribution (OOD) \textit{natural} examples. ~\citeauthor{liu2022sample} (\citeyear{liu2022sample}) find that prompt-based few-shot learning approaches lead to \textit{more robust models} than their fully fine-tuned counterparts. ~\citeauthor{awadalla2022exploring} arrive at the same finding in the specific context of Question Answering tasks. However, since both works focus on out-of-distribution samples that are considered likely and natural, it is unclear if their findings also hold for samples that attackers adversarially perturb. Consequently, we specifically focus on the adversarial robustness of data-efficient learning for NLU. Our findings show that, contrary to the trends observed for OOD samples in prior works, in-domain performance is not a good predictor of adversarial robustness of prompt-based few-shot learning approaches compared to fully supervised approaches. In other words, fully supervised models demonstrate a lesser relative drop in adversarial performance with respect to in-domain performance than prompt-based few-shot approaches. However, when strategies such as \textit{(a)} using unlabeled data and \textit{(b)} ensembling over models trained with multiple prompts are adopted, the resultant models demonstrate better adversarial robustness than fully fine-tuned models.  

\section{Experimental Setup} 
\noindent\textbf{Few-shot Learning (FSL) Methods using Prompting}: 
We evaluate four different FSL methods that are commonly used for natural language understanding tasks: Classic fine-tuning~\cite{devlin2018bert}, LM-BFF~\cite{gao2020making}, PET, and iPET ~\cite{schick2020s, schick2021exploiting}. Together, these approaches cover three primary settings in state-of-the-art prompt-based FSL methods, namely, (i) no use of unlabeled data for training, (ii) use of unlabeled data, and (ii) using ensembles of models trained with different prompts. We consider fine-tuning with fully labeled data to give the ceiling performance and contrast the capabilities of the FSL methods. Below, we briefly describe the FSL methods and explain our rationale for considering them in our study.

\vspace{0.05in}
\noindent\textbf{1. Classic-FT}: We use the \texttt{[CLS]} token representation from the encoder with a softmax classifier on top and train the model end-to-end on a few labeled examples (no unlabeled data). 

\vspace{0.05in}
\noindent\textbf{2. LM-BFF}: ~\citeauthor{gao2020making} (\citeyear{gao2020making}) proposed few-shot fine-tuning with prompting using demonstrations. Their approach for FSL involves \textit{concatenating} the input example, which is modified to follow the prompting template with a \texttt{[MASK]} in place of the verbalizer, with semantically similar examples (i.e., demonstrations) from the few-shot training set. Concatenating one demonstration per class with the input example enables overcoming the long-context problem of GPT-3's in-context learning. During inference, LM-BFF ensembles the predictions made by concatenating the input example with all demonstrations from the few-shot training set. LM-BFF does not use unlabeled data for training. 

\vspace{0.05in}
\noindent\textbf{3. PET}: Pattern-Exploiting Training (PET)~\cite{schick2020s} is a simple prompt-based few-shot fine-tuning approach where the training examples are converted into templates, and the \texttt{[MASK]} tokens are used to predict the verbalizer, which indicates the output label. To understand the role of using multiple prompts in robustness, we use PET to fine-tune models with different template-verbalizer pairs and ensemble their predictions during inference. PET does not use demonstrations or unlabeled data.

\vspace{0.05in}
\noindent\textbf{4. iPET}: iPET~\cite{schick2020s, schick2021exploiting} involves self-training and leverages unlabeled data during fine-tuning. It iteratively uses PET to produce multiple generations, assigning pseudo-labels to unlabeled data at the end of each generation stage. This pseudo-labeled data from a previously fine-tuned model is then used along with the few-shot training data to update the model in the subsequent generation stage. iPET uses unlabeled data and allows us to understand its impact on adversarial robustness.

\vspace{0.05in}
\noindent\textbf{GLUE and AdvGLUE Benchmarks}:
We train the above FSL methods on $6$ GLUE~\cite{wang2018glue} tasks that also have a corresponding adversarial counterpart in the Adversarial-GLUE (AdvGLUE) benchmark~\cite{zheng2021fewnlu}, namely, SST-2~\cite{socher2013recursive}, QQP\footnote{\url{https://www.quora.com/profile/Ricky-Riche-2/First-Quora-Dataset-Release-Question-Pairs}}, MNLI-m, MNLI-mm~\cite{williams2017broad}, RTE~\cite{dagan2006pascal, haim2006second, giampiccolo2007third, bentivogli2009fifth}, and QNLI~\cite{rajpurkar2016squad}. These tasks consider sentences or sentence pairs as input. The existence of a corresponding adversarial counterpart enables systematic assessment of these FSL methods trained on the original in-domain datasets. The AdvGLUE corpus comprises task-specific adversarial examples obtained using 14 textual adversarial attack methods. Recall that the adversarial attack methods cover word-level and sentence-level perturbations, as well as human-crafted examples. Since \citeauthor{wang2021adversarial} (\citeyear{wang2021adversarial}) find that, in certain cases, as many as ~90\% adversarial examples constructed using automated methods are invalid, they perform human validations to ensure that only valid adversarial perturbations are included in this benchmark dataset. 

\subsection{Implementation Details}
\begin{table*}[]
    \centering
    \begin{tabular}{c| l | l}
    \toprule
        \textbf{Task} & \textbf{Template} & \textbf{Verbalizer}\vspace{1mm}\\\midrule
        SST-2 & $<S_1>$ {It was [MASK] .} & positive: great, negative: terrible \\
        QQP & $<S_1>$ {[MASK] ,} $<S_2>$  & equivalent: Yes, not\_equivalent: No\\
        MNLI & $<S_1>$ ? {[MASK] ,} $<S_2>$ & entailment: Yes, neutral: Maybe, contradiction: No\\
        RTE & $<S_1>$ ? {[MASK] ,} $<S_2>$ & entailment: Yes, not\_entailment: No\\
        QNLI & $<S_1>$ ? {[MASK] ,} $<S_2>$ & entailment: Yes, not\_entailment: No \vspace{1mm}\\\bottomrule
    \end{tabular}
    \caption{Prompts used in this study adopted from ~\citeauthor{gao2020making} (\citeyear{gao2020making}). $<S_1>$ and $<S_2>$ are the input sentences.}
    \label{tab:prompts}
\end{table*}

\begin{table}[!t]
    \centering
    \resizebox{1.0\columnwidth}{!}{%
    \begin{tabular}{c| l | l}
    \toprule
        \textbf{Task} & \textbf{Template} & \textbf{Verbalizer}\vspace{1mm}\\\midrule
        \multirow{4}{*}{SST-2} & {It was [MASK] .} $<S_1>$  & bad / good \\ 
        & $<S_1>$ {. All in all, it was [MASK] . }& bad / good \\
         &  {Just [MASK] !} $<S_1>$ & bad / good \\         
          & $<S_1>$ { In summary, the movie was [MASK] . }& bad / good \\\midrule
         \multirow{4}{*}{QQP} & $<S_1>$ {[MASK] ,} $<S_2>$ & No / Yes \\
         & $<S_1>$ {[MASK], I want to know} $<S_2>$ & No / Yes \\
         & $<S_1>$ {[MASK], but} $<S_2>$ & No / Yes \\
         & $<S_1>$ {[MASK], please ,} $<S_2>$ & No / Yes \\\midrule
         \multirow{4}{*}{MNLI} & $<S_1>$ ? {[MASK] ,} $<S_2>$ & Wrong/Right/Maybe\\
         & $<S_1>$ ? {[MASK] ,} $<S_2>$ & No/Yes/Maybe\\
         & " $<S_1>$" ? {[MASK] ,} " $<S_2>$ "& No/Yes/Maybe\\
         & " $<S_1>$" ? {[MASK] ,} " $<S_2>$ " & Wrong/Right/Maybe\\\midrule
         \multirow{4}{*}{RTE} & " $<S_2>$ " ? {[MASK] ,} " $<S_1>$ "& No/Yes \\
         & $<S_2>$ ? {[MASK] ,} " $<S_1>$ "& No/Yes \\
         & " $<S_1>$ " ? {[MASK] .} $<S_2>$ & No/Yes \\
         & $<S_1>$ ? {[MASK] .} $<S_1>$ & No/Yes \\\midrule
         \multirow{4}{*}{QNLI} & $<S_1>$ ? {[MASK] ,} $<S_2>$ &  No/Yes\\
         & $<S_1>$ ? {[MASK] ,} $<S_2>$ &  Wrong/Right\\
         & " $<S_1>$" ? {[MASK] ,} " $<S_2>$ "& No/Yes\\
         & " $<S_1>$" ? {[MASK] ,} " $<S_2>$ " & Wrong/Right No/Yes\\\bottomrule
    \end{tabular}
    }
    \caption{Manual template and verbalizer pairs used for PET. $<S_1>$ and $<S_2>$ are the input sentences.}
    \label{tab:PETprompts}
\end{table}

\vspace{0.05in}
\noindent\textbf{Evaluation Protocol}: Our experimental setup involves taking each FSL method described earlier and training the model using $K$ randomly sampled examples \textit{per class} from the original in-domain train set. We then evaluate the performance of the resulting models on two evaluation sets for each task: the original GLUE evaluation set (in-domain) and the corresponding adversarial version in AdvGLUE. For our main results, we use  $K = 64$ examples per class. We also perform ablations by varying $K \in \{16, 32, 64, 128, 256\}$. For each of the aforementioned FSL approaches, we use ALBERT-xxlarge-v2~\cite{lan2019albert} as the pre-trained language model for our experiments. We conduct ablations by varying the ALBERT encoder size to be \texttt{base} (12M),  \texttt{large} (18M), \texttt{xlarge} (60M), \texttt{xxlarge} (235M), and encoder type as BERT~\cite{devlin2018bert}, RoBERTa~\cite{liu2019roberta}, and ALBERT~\cite{lan2019albert}. We quantify the performance of these models using Accuracy values (and $F_1$ score for QQP). We also quantify the gap between in-domain and adversarial performance using a relative percent drop in Accuracy/F1 scores. 

\vspace{0.05in}
\noindent\textbf{Prompting}: Since LM-BFF, PET, and iPET are prompt-based fine-tuning methods, an important consideration while comparing their performance is to use comparable prompts. A prompt comprises of two parts: a \textit{template} phrase that is appended to the input and a \textit{verbalizer} that maps to the output label. For instance, for a sentence $s_1 =$ \textit{``this was probably the best pizza in entire city''}, the prompt $p =$ \textit{``It was [MASK]''} is concatenated (i.e., $s_1 \oplus p$), and the model is trained to predict the words ``great'' and ``terrible'' that map to the sentiment labels `positive' and `negative', respectively. We use the prompts (i.e., templates as well as verbalizers) identified by ~\citeauthor{gao2020making} (\citeyear{gao2020making}) for all the approaches; we list them in Table \ref{tab:prompts}. Experiments with PET require additional prompts to isolate the effect of ensembling predictions of models trained using different prompts; we list the prompts used for training PET in Table \ref{tab:PETprompts}.

\subsection{Method-specific Design Choices}
As mentioned earlier, for our main experiments, we used the \texttt{xxlarge} variant of the ALBERT encoder (Albert-xxlarge-v2) as the MLM encoder. All our experiments were conducted using a single GPU with 48GB RAM (NVIDIA Quadro RTX 8000). 
To eliminate the need for an extensive hyper-parameter search, for each of the prompting methods, unless otherwise stated, we use the same set of hyperparameters as recommended in ~\citeauthor{gao2020making} (\citeyear{gao2020making}); most notably,  batch size of 8, learning rate set to $10^-5$, and max sequence length of 256. 

\vspace{0.05in}
\noindent\textbf{LM-BFF Considerations}: We used demonstrations along with manual prompts listed in Table \ref{tab:prompts}. We do not use automatic prompt generation as specifying a manual prompt allows controlled comparison across different prompting methods, some of which can only use manually-specified prompts.  Furthermore, automated prompts increase the training cost. For demonstrations, we concatenate one semantically similar example per class to the input example during the training phase. During inference, for each test example, we ensemble the predictions over different possible sets of demonstrations. To control for the sensitivity of prompting to the selected sample, we perform random sampling and subsequent training of LM-BFF for $N = 5$ times and 1000 training steps, for each task.

\vspace{0.05in}
\noindent\textbf{iPET Considerations}: For iPET, we train the models on two randomly sampled data folds, with each fold having K= 64 examples per class, for a total of 3 generations and 250 training steps to speed up the training process. The unlabeled dataset size is limited to 500 examples with a scale factor of 3 (i.e., in every generation, the total training dataset size is increased by a factor of 3). In the subsequent generation stage, the model trained on one data fold is used to generate the pseudo-labeled training set for the model trained on the other fold. We evaluate the models obtained after the final generation. 

\vspace{0.05in}
\noindent\textbf{PET Considerations}: We train the model on four different sets of manual template-verbalizer pairs for 250 training steps. The manual template-verbalizer pairs used for different tasks are listed in Table \ref{tab:PETprompts}. We arrive at these prompts based on the templates proposed for similar tasks by ~\citeauthor{schick2020s} (\citeyear{schick2020s}), and by using the prompts specified for LM-BFF by ~\citeauthor{gao2020making} (\citeyear{gao2020making}). During inference, we evaluate the ensemble of models trained on all the different prompts. 

\section{Results}

\begin{table*}[!t]
    \resizebox{1.0\textwidth}{!}{
    \begin{tabular}{c  c  c  c  c  c  c  c c }
    \toprule
        \textbf{Method}  & \textbf{Setting} & \textbf{Average} $\uparrow$ & \multicolumn{6}{c}{\textbf{Tasks}}\\
         & & & \textbf{SST-2 } $\uparrow$  & \textbf{QQP} $\uparrow$ & \textbf{MNLI-m} $\uparrow$ & \textbf{MNLI-mm} $\uparrow$ & \textbf{RTE} $\uparrow$ & \textbf{QNLI} $\uparrow$ \\\midrule
        \multirow{1}{*}{Full FT} & Org & $91.7$ &  $95.2$ & $92.3/89.5$ & $89.3$ & $89.9$ & $88.4$ & $95.3$ \\
         & Adv & $59.3$ & $66.8$ & $56.4$ / $32.4$ & $51.8$ & $44.2$ & $73.0$ & $63.8$ \\\midrule
         \multirow{2}{*}{Classic FT} & Org & $66.2$ & $85.6$ \small{$(\pm 3.1)$} & $75.0$ \small{$(\pm 3.0)$} / $68.3$ \small{$(\pm 6.0)$} & $52.3$ \small{$(\pm 5.0)$} & $53.5$ \small{$(\pm 4.8)$} & $56.9$ \small{$(\pm 1.6)$}& $76.8$ \small{$(\pm 3.2)$}  \\
         & Adv & $50.9$ & $56.2$ \small{$(\pm 2.2)$} & $57.2$ \small{$(\pm 8.8)$} / $52.9$ \small{$(\pm 9.8)$} & $37.7$ \small{$(\pm 9.3)$} & $41.6$ \small{$(\pm 9.3)$} & $53.3 $ \small{$(\pm1.6)$} & $59.6$ \small{$(\pm5.6)$}\\\midrule
        \multirow{2}{*}{LM-BFF} & Org & $81.4$ & $94.0$ \small{($\pm 0.4$)} & $80.1$ \small{$\pm 0.7$} / $75.6$ \small{$(\pm 0.9)$}& $76.7$ \small{$(\pm 1.2)$} & $78.3$ \small{$(\pm 1.3)$} & $78.1$ \small{$(\pm 2.5)$}& $81.4$ \small{$(\pm 2.0)$}\\
         & Adv & $51.3$ & $54.1$ \small{$(\pm 0.9)$} & $46.2$ \small{$(\pm 6.4)$} / $46.1$ \small{$(\pm 6.1)$} & $47.1$ \small{$(\pm 1.5)$} & $40.1$ \small{$(\pm 3.2)$} & $58.8$ \small{$(\pm 3.8)$}& $61.5$ \small{$(\pm 4.2)$}\\\midrule
        \multirow{2}{*}{iPET} & Org & $80.8$ & 93.4 \small{$(\pm 0.4)$}& 79.4 \small{$(\pm 0.4)$} / 74.5 \small{$(\pm 0.8)$}& 76.1 \small{$(\pm 0.9)$}  & 77.3 \small{$(\pm 0.6)$} & 74.2 \small{$(\pm 0.2)$} & 84.6 \small{$(\pm 1.3)$}  \\
         & Adv & $58.1$ & 65.9 \small{$(\pm 1.4)$} & 59.6 \small{$(\pm 9.9)$} / 59.4 \small{$(\pm 8.9)$} & 60.3 \small{$(\pm 1.2)$} & 47.2 \small{$(\pm 1.3 )$}& 58.1 \small{$(\pm 5.2)$} & 57.4 \small{$(\pm 3.8)$} \\\midrule
         \multirow{2}{*}{PET} & Org & $78.6$ & 93.4  \small{$(\pm 0.5 )$}  & 73.7 \small{$(\pm 4.5 )$} / 68.6 \small{$(\pm 2.4)$}  & 74.6 \small{$(\pm 3.8)$} & 75.7 \small{$(\pm 3.6)$}& 72.5 \small{$(\pm 7.2)$}& 81.6 \small{$(\pm 1.5)$}  \\
         & Adv & $57.2$ &  61.7 \small{$(\pm 1.7 )$}& 59.3 \small{$(\pm 1.6 )$} / 55.2 \small{$(\pm 5.2 )$}& 55.6 \small{$(\pm 4.5)$} & 44.8 \small{$(\pm 5.8)$} & 54.0 \small{$(\pm 4.1)$} &  67.9 \small{$(\pm 1.6 )$} \\\bottomrule
    \end{tabular}
    }
    \caption{Performance comparison of different methods on in-domain evaluation sets of GLUE (Org) and Adversarial GLUE (Adv) benchmarks. \texttt{ALBERT-xxlarge-v2} is used as the large pre-trained language model. We report the average and standard deviation in the accuracy values of $5$ different runs. For these results, we set $K = 64$. $\uparrow$ denotes that a higher value indicates better performance. Average is the average accuracy across all tasks. }
    \label{tab:main_results}
\end{table*}

\begin{table*}[!t]
    \centering
    \resizebox{0.83\textwidth}{!}{
    \begin{tabular}{c  c  c  c  c  c  c c }
    \toprule
        \textbf{Method}  & \textbf{Average} & \multicolumn{6}{c}{\textbf{Tasks}}\\
         & \textbf{Drop} $\downarrow$ & \textbf{SST-2} $\downarrow$ & \textbf{QQP} $\downarrow$ & \textbf{MNLI-m} $\downarrow$ & \textbf{MNLI-mm} $\downarrow$ & \textbf{RTE} $\downarrow$ & \textbf{QNLI} $\downarrow$\\\midrule
        \multirow{1}{*}{Full FT}
         & $35.3$ & $30.4$ & $38.7$ / $63.7$ & $42.3$ & $50.8$ & $15.7$ & $32.2$ \\\midrule
         \multirow{1}{*}{Classic FT}
         & $23.1$ & $34.3$ & $23.7$ / $22.5$ & $27.9$ & $22.2$  & $06.3$  & $22.4$ \\\midrule
        \multirow{1}{*}{LM-BFF} 
         & $36.9$ & $42.4$  & $42.3$  / $39.0$ & $38.6$  & $48.8$  & $24.7$ & $24.4$ \\\midrule
        \multirow{1}{*}{iPET} & $28.1$ & $29.4$ & $24.9$ / $20.2$ & $20.8$ & $38.9$ & $21.7$ & $32.1$ \\\midrule
        \multirow{1}{*}{PET} & $27.2$ & $33.9$ & $$ 19.5 / 18.9 $$ & $24.6$ & $40.8$ & $25.5$ & $16.8$ \\\bottomrule
    \end{tabular}
    }
    \caption{Relative performance drop between in-domain evaluation set of GLUE (Org) and AdvGLUE (Adv) test set of different methods given by $(Org-Adv)/Org \times 100$. \texttt{ALBERT-xxlarge-v2} is used as the large pre-trained language model. We report the average and standard deviation of accuracy values of $5$ different runs. For these results, we set $K = 64$. $\downarrow$ denotes that a lower value indicates better performance. Average Drop is the relative drop in average accuracy values computed in Table \ref{tab:main_results}.}
    \label{tab:relative_drops}
\end{table*}

\noindent\textbf{Robustness of FSL Methods}:
In Table \ref{tab:main_results} we show the performance of few-shot learning methods on in-domain GLUE and AdvGLUE evaluation sets using accuracy values (along with $F_1$ score for QQP). In Table \ref{tab:relative_drops}, we present the relative decrease in performance on the AdvGLUE benchmark with respect to the performance on the GLUE evaluation set. This relative drop is critical to quantify as our focus is on understanding the \textit{surprise} in terms of a fine-tuned model's performance on an adversarial test set with respect to its performance on the in-domain evaluation set. In other words, the relative drop answers the following question: \textit{is the classification performance on the in-domain evaluation set a reliable estimate of performance in the face of adversarial inputs}? 

We find that  classic fine-tuning experiences a lesser relative drop in performance (i.e. it is more robust) in 5 out of 6 GLUE tasks, when compared to LM-BFF. However, as expected, ClassicFT also leads to subpar performance on the original GLUE evaluation set, which limits its usability as an efficient FSL technique. While LM-BFF provides good few-shot performance on the GLUE benchmark, it demonstrates poorer adversarial robustness than full fine-tuning in 4 out of 6 tasks. Moving to iPET, we observe that including unlabeled data with prompt-based FSL leads to a lesser relative performance drop in 5 out of 6 tasks when compared to full fine-tuning. Finally, the inclusion of multiple prompts in PET demonstrates a similar effect – that is, a lesser relative performance drop in 4 out of 6 tasks over full fine-tuning. Collectively, these trends demonstrate the benefits of using unlabeled data and ensembling towards greater adversarial robustness of prompt-based FSL. Note that the trends described using the observed relative performance drops on the majority of tasks are the same as the trends observed with average accuracy values across tasks (i.e., `Average' \& `Average Drop' in Tables \ref{tab:main_results} \& \ref{tab:relative_drops}). 

Overall, our experiments demonstrate that prompt-based FSL methods that use only demonstrations (i.e., LM-BFF) severely lag in terms of their adversarial robustness, performing worse than simple classic fine-tuning (i.e., ClassicFT) with the same number of examples. However, leveraging unlabeled data and ensembles trained with different prompts separately (i.e., via iPET and PET, respectively) improve the adversarial robustness of prompt-based FSL over fully supervised fine-tuning (i.e., FullFT). We briefly discuss the role of these modeling choices when used with prompting in improving the adversarial performance relative to in-domain performance. 

iPET uses unlabeled training data during fine-tuning by iteratively training the models on pseudo-labels generated by previous models. In the process, the model is exposed to more diverse samples of the data than simple prompt-based learning (i.e., LM-BFF in our case). \citeauthor{alayrac2019labels} (\citeyear{alayrac2019labels}) show that unlabeled data is an effective alternative to labeled data for training  adversarially robust models. Our findings in the context of prompting language models for few-shot learning support their original claims made in the context of image classification tasks. 
Additionally, prior work has shown that prompt-based few-shot performance is sensitive to the prompts used for training and has used that observation to automatically find prompts that provide maximum performance on in-domain evaluation sets~\cite{gao2020making}. Similarly, ensembling  predictions of models trained using multiple prompts is also found to be better than relying on a single prompt~\cite{zheng2021fewnlu}. From our results, we observe that ensembling also helps overcome the sensitivity of a single model to variations in input data, especially adversarial variations. 

\begin{table}[!t]
    \centering
    \resizebox{0.8\linewidth}{!}{
    \begin{tabular}{c  c  c  c }
    \toprule
        \textbf{K} &  \textbf{Setting} & \multicolumn{2}{c}{\textbf{Tasks}}\\
         & & \textbf{SST-2} & \textbf{MNLI-m} \\\midrule
        \multirow{1}{*}{16}
         & Org & 92.6 $(1.2)$& 69.1 (2.0)\\
         & Adv & 56.5 $(5.0)$& 49.1 (4.8)\\\midrule
         \multirow{1}{*}{32}
         & Org & 93.2 $(1.0)$& 75.2 (1.3)\\
         & Adv & 55.3 $(3.0)$& 49.9 (3.2)\\\midrule
         \multirow{1}{*}{64}
         & Org & 94.0 $(0.4)$& 76.7 (1.2)\\
         & Adv & 54.1 $(0.9)$& 47.1 (1.5)\\\midrule
         \multirow{1}{*}{128}
         & Org & 94.2 (0.3)& 80.8 (0.4)\\
         & Adv & 58.8 (2.3)& 51.7 (3.4)\\\midrule
         \multirow{1}{*}{256}
         & Org & 94.7 (0.3)& 83.2 (0.7)\\
         & Adv & 63.1 (2.9)& 53.6 (1.5)\\\bottomrule
         \end{tabular}
         }
    \caption{ Effect of varying the number of few-shot labeled examples ($K$) on adversarial (Adv) and in-domain (Org) performance for SST-2 and MNLI-m tasks.}
    \label{tab:ablation_k}
\end{table}

\vspace{0.05in}
\noindent\textbf{Effect of the number of few-shot examples, the encoder size and type}:
To isolate the effect of the number of few-shot examples, the encoder size (in terms of the number of learnable parameters), and the encoder type, we fix the FSL method to LM-BFF and vary these factors one at a time. Additionally, we conduct ablation experiments on two representative tasks, SST-2 and MNLI-m.

\begin{figure}[!t]
    \centering
    \includegraphics[width = 1.0\linewidth]{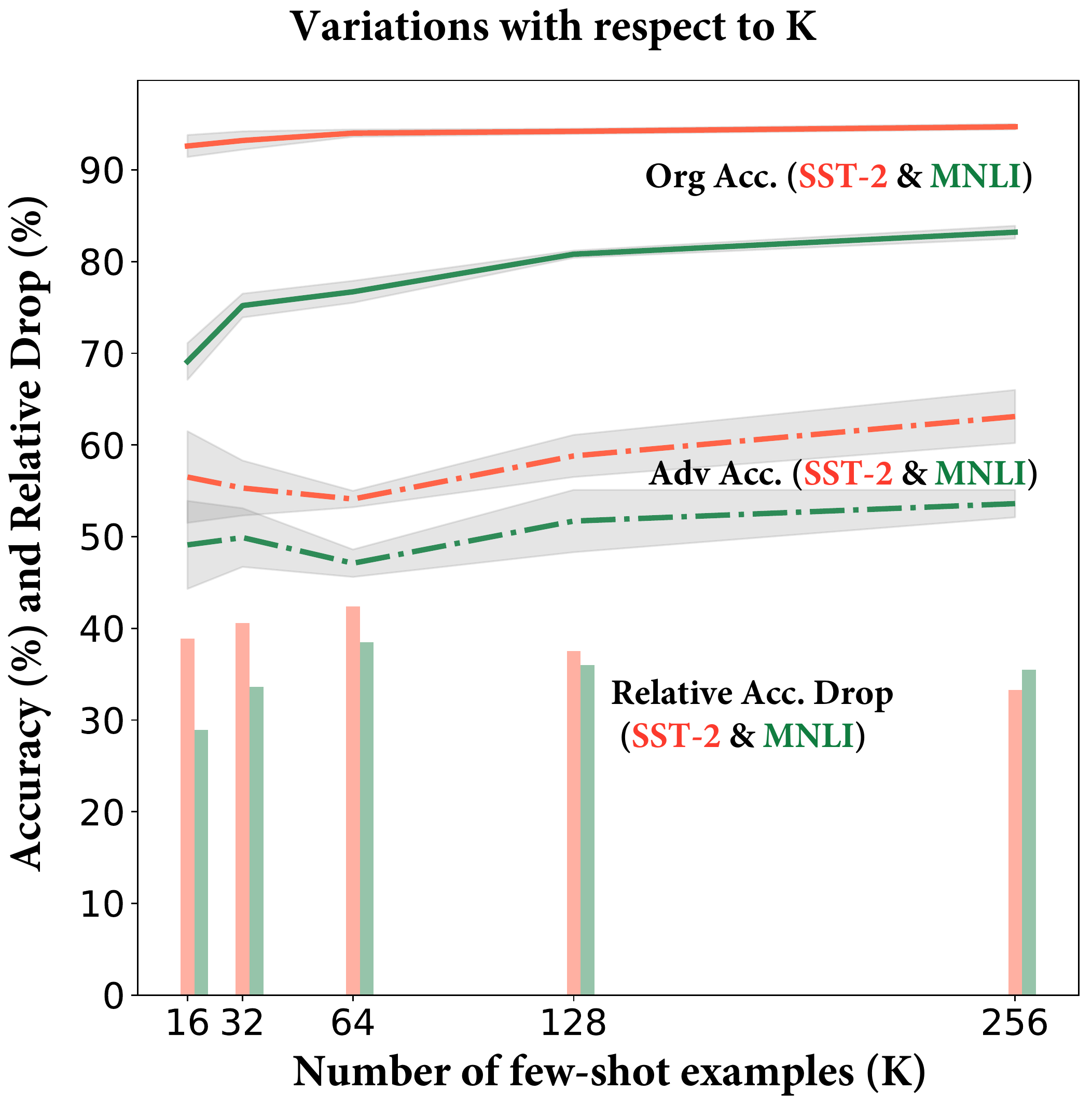}
    \caption{Variation in in-domain (Org; solid) and adversarial performance (Adv; dash-dotted) in terms of accuracy with respect to the number of few-shot examples $K$, for \textcolor{red}{SST-2} and \textcolor{darkgreen}{MNLI-m}. We also show the variation in the relative percentage drop in accuracy given by $(org - adv)/org\ \%$ using bar charts.}
    \label{fig:k_variation_figure}
\end{figure}

Table \ref{tab:ablation_k} and Figure \ref{fig:k_variation_figure} show that increasing the number of examples for few-shot learning improves performance on both in-domain GLUE and Adversarial GLUE evaluation sets. Interestingly, the relative performance drop on the adversarial set with respect to the in-domain set diminishes slightly, indicating that more examples are helpful in bridging the gap between in-domain performance and adversarial robustness. The results are consistent across both tasks. Since the essence of FSL methods is in learning effectively with little data, this observation provides further evidence that current few-shot models demonstrate a trade-off between in-domain performance and adversarial robustness. Another key aspect of resource-efficient learning (besides data-efficient learning) is learning with a limited number of parameters. Next, we investigate the effect of model size on the model’s adversarial robustness.

\begin{table}[!t]
    \centering
    \resizebox{\linewidth}{!}{
    \begin{tabular}{c  c  c  c  c}
    \toprule
        \textbf{Version} & \textbf{Size} &  \textbf{Setting} & \multicolumn{2}{c}{\textbf{Tasks}}\\
         & & & \textbf{SST-2} & \textbf{MNLI-m} \\\midrule
        \multirow{1}{*}{base} 
         & 12M & Org & 85.6 (0.7)& 52.5 (2.5) \\
         & & Adv & 34.2 (4.0) & 32.9 (4.6)\\\midrule
         \multirow{1}{*}{large}
         & 18M & Org & 88.0 (0.7)& 61.2 (0.9) \\
         & & Adv & 36.4 (3.8)& 39.5 (2.8) \\\midrule
         \multirow{1}{*}{xlarge}
         & 60M & Org & 89.3 (0.8)& 67.4 (2.9) \\
         & & Adv & 45.7 (4.6)& 39.3 (4.8)\\\midrule
         \multirow{1}{*}{xxlarge}
         & 235M & Org & 94.0 (0.4)& 76.7 (1.2)\\
         & & Adv & 54.1 (0.9)& 47.1 (1.5) \\\bottomrule
         \end{tabular}
         }
    \caption{ Effect of variation in encoder size on in-domain (Org) and adversarial (Adv) performance for SST-2 and MNLI-m tasks.}
    \label{tab:ablation_size}
\end{table}

\begin{figure}[!t]
    \centering
    \includegraphics[width = 1.0\linewidth]{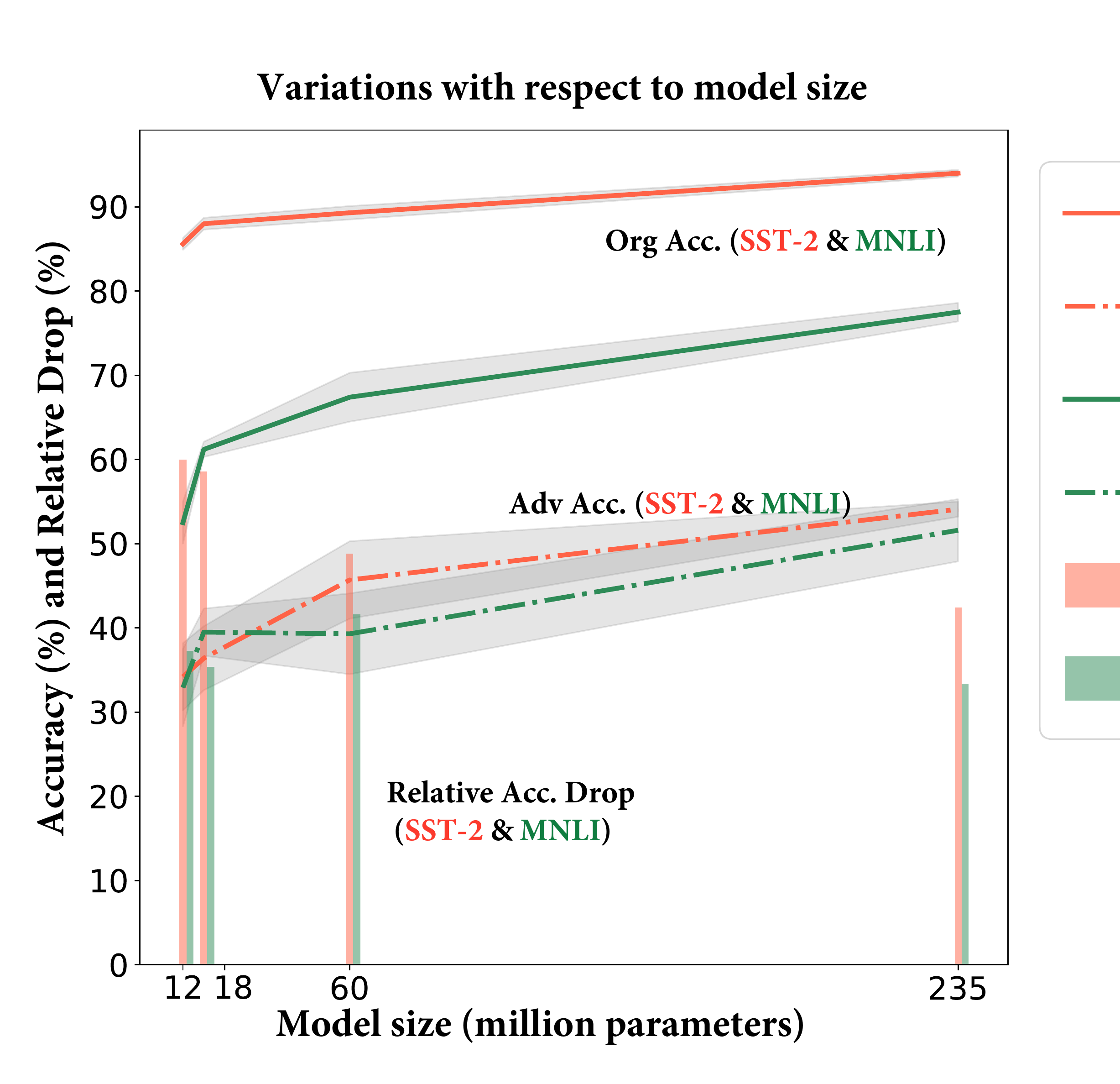}
    \caption{Variation in in-domain (Org; solid) and adversarial performance (Adv; dash-dotted) in terms of accuracy with respect to the model size, for \textcolor{red}{SST-2} and \textcolor{darkgreen}{MNLI-m}. We also show the variation in the relative percentage drop in accuracy given by $(org - adv)/org\ \%$) using bar charts.}
    \label{fig:model_variation_figure}
\end{figure}

In Table \ref{tab:ablation_size} and Figure \ref{fig:model_variation_figure}, we present the results by varying the encoder size of the ALBERT model used in LM-BFF, while keeping the number of examples used for training as 64. Results show that as the size of the encoder increases in the number of learnable parameters, the performance on both evaluation set increases, and the gap between in-domain performance and adversarial robustness decreases. The performance gap is drastic in smaller encoders like base (12M) and large (18M). The observed results are consistent across both tasks.

\begin{table}[!t]
    \centering
    \resizebox{\linewidth}{!}{
    \begin{tabular}{c  c  c  c  c}
    \toprule
        \textbf{Encoder} & \textbf{Size}  & \textbf{Setting} & \multicolumn{2}{c}{\textbf{Tasks}}\\\toprule
         & & & \textbf{SST-2} & \textbf{MNLI-m} \\\midrule
        \multirow{1}{*}{BERT-large-uncased}
         & 334M & Org & 89.8 (0.9)& 57.8 (0.3)\\
         & & Adv & 29.9 (2.8)& 35.5 (5.0)\\\midrule
         \multirow{1}{*}{RoBERTa-large}
         & 355M & Org & 93.5 (0.5)& 77.5 (0.6) \\
         & & Adv & 58.8 (4.2)& 53.5 (2.1)\\\midrule
        \multirow{1}{*}{ALBERT-xxlarge-v2}
         & 235M & Org & 94.4 (0.4)& 77.5 (1.1) \\
         & & Adv & 54.1 (0.9)& 51.6 (3.7)\\\bottomrule
         \end{tabular}
         }
    \caption{Effect on in-domain (Org) and adversarial (Adv) performance with variation in encoder type. We experiment with three different encoders (BERT, RoBERTa, and ALBERT) of comparable sizes ($\sim10^8$).}
    \label{tab:ablation_type}
\end{table}

Finally, we again keep the number of examples as 64 and vary the encoder type to be one of the three widely-used large language models: BERT, RoBERTa, and ALBERT. To control the effect of different encoder sizes, we keep the encoder parameters in a similar range ($10^8$). We notice that RoBERTa encoder is the most effective in balancing the trade-off between in-domain performance and adversarial robustness. ALBERT demonstrates on-par in-domain performance but lags slightly in adversarial robustness. This observation could be attributed to RoBERTa having $34\%$ more parameters than ALBERT. BERT demonstrates the worst trade-off between in-domain performance and adversarial robustness. Since the fine-tuning strategy adopted with these models is the same, the observed trends could be attributed to the pre-training approach for these encoders. For instance, whole-world masking (used for pre-training RoBERTa) is found to be more adversarially robust than masked language modeling (used for pre-training BERT)~\cite{dong2021should}, indicating that that the former leads to adversarially reliable textual representations that also model syntax and sentence structure better.

\section{Discussion and Conclusion}
\noindent\textbf{Adversarial robustness versus OOD robustness}: Recent prior work by ~\citeauthor{awadalla2022exploring} and ~\cite{liu2022sample} explore the out-of-distribution (OOD) robustness of prompt-based FSL methods and find that prompting leads to more robust models than fully fine-tuned models. However, we find that these results do not extend to adversarial robustness where the examples are crafted by adversaries (either humans or machines) to fool the models. While prompting methods can improve the end-user experience with language technologies by performing better on OOD samples, they also leave such technologies more vulnerable to adversarial attacks by malicious agents. We encourage the community to consider robustness along both of these axes while developing and evaluating future prompting methods. 

Considering adversarial robustness is especially important because prompt-based few-shot learning has recently found applications in societal tasks like hate speech detection~\cite{wang2021entailment}, toxicity detection~\cite{wang2022toxicity}, and author profiling~\cite{chinea2022zero}. Prompting allows us to leverage ever-evolving data in the real world with limited annotation efforts. However, prompt-based FSL methods can be manipulated by well-coordinated adversaries using carefully crafted inputs on social platforms, and the end-users could be exposed to incorrectly filtered, and potentially harmful, content by these language technologies. Therefore, we recommend researchers and practitioners exercise caution while applying prompt-based few-shot learning to societal tasks.

\noindent\textbf{Costs of obvious solutions}: In our work, we have isolated different factors that impact the adversarial robustness of prompt-based FSL. However, each of these factors is associated with additional costs. Reliance on unlabeled data during fine-tuning requires curation, albeit no annotation. Few-shot learning with multiple prompts incurs additional training costs and inference time as predictions from multiple models are ensembled. Increasing the number of few-shot examples goes against the premise of few-shot learning. Similarly, increasing model size leads to models that are difficult to deploy in practice. These pose new challenges for NLP researchers and practitioners as adversarial robustness is a critical constraint along with other constraints like in-domain performance, OOD robustness, data, energy, \& parameter efficiency.

\section{Limitations and Broader Perspective}

\vspace{0.01in}\noindent\textit{Limitations and Future Work}: As the first study to assess the adversarial robustness of prompt-based FSL methods, we focus on representative methods that cover different design choices. Future work could expand the set of prompt-based FSL methods considered in this study. Our broader goal is to encourage systematic evaluation of adversarial robustness for all prompt-based FSL methods. Furthermore, we do not perform extensive hyperparameter tuning for the methods considered in this work. It is worth noting that ``true'' few-shot learning setting has been argued not to involve any development set (as that would involve collecting more labeled data)~\cite{perez2021true, schick2022true}. To this end, we use the hyper-parameters reported by the original authors of these methods. Future work could explore settings where access to a limited development set is assumed for exhaustive hyperparameter tuning. Finally, for adversarial evaluation of prompt-based FSL approaches, we utilize a pre-constructed dataset --- AdvGLUE ~\cite{wang2021adversarial}. Since these examples are pre-constructed, they do not have access to the gradients of the specific victim models under investigation. Nonetheless, the AdvGLUE benchmark offers a foundation for understanding vulnerabilities in large-scale language models under various adversarial scenarios. This standardized dataset enables fair comparison and mitigates issues with invalid perturbations. For instance, ~\citeauthor{wang2021adversarial} (\citeyear{wang2021adversarial}) found that over 90\% of adversarial perturbations generated using the gradients of victim models for NLP tasks are invalid. Therefore, using AdvGLUE ensures adversarial evaluation on high-quality, human-verified data. Future work could extend the study by considering adversarial examples generated using the gradients of victim models and validating them for correctness.

\vspace{0.01in}\noindent\textit{Broader Social Impact}: 
The authors do not foresee any negative social impacts of this work. We believe systematic and preemptive evaluation of the robustness of language technologies against potential adversarial attacks will help develop more safe and secure systems. We release the code for our experiments to aid reproducibility and promote future research on this topic.

\vspace{0.01in}\noindent\textit{Datasets}: The datasets used for this study are publicly available and were curated by previous research;  no new data was collected for this study. We abide by the terms of use of the benchmarks as well as the individual datasets. 

\section{Acknowledgements}
This research/material is based upon work supported in part by NSF grants CNS-2154118, IIS-2027689, ITE-2137724, ITE-2230692, CNS-2239879, Defense Advanced Research
Projects Agency (DARPA) under Agreement No.
HR00112290102 (subcontract No. PO70745), and funding from Microsoft, Google, and Adobe Inc. GV is partly supported by the Snap Research Fellowship. Any opinions, findings, and conclusions or recommendations expressed in this material are those of the author(s) and do not necessarily reflect the position or policy
of DARPA, DoD, SRI International, NSF and no official endorsement should be inferred. We thank the anonymous reviewers for their constructive comments.

\bibliography{anthology,custom}

\begin{thebibliography}{45}
\expandafter\ifx\csname natexlab\endcsname\relax\def\natexlab#1{#1}\fi

\bibitem[{Alayrac et~al.(2019)Alayrac, Uesato, Huang, Fawzi, Stanforth, and
  Kohli}]{alayrac2019labels}
Jean-Baptiste Alayrac, Jonathan Uesato, Po-Sen Huang, Alhussein Fawzi, Robert
  Stanforth, and Pushmeet Kohli. 2019.
\newblock Are labels required for improving adversarial robustness?
\newblock \emph{Advances in Neural Information Processing Systems}, 32.

\bibitem[{Awadalla et~al.(2022)Awadalla, Wortsman, Ilharco, Min, Magnusson,
  Hajishirzi, and Schmidt}]{awadalla2022exploring}
Anas Awadalla, Mitchell Wortsman, Gabriel Ilharco, Sewon Min, Ian Magnusson,
  Hannaneh Hajishirzi, and Ludwig Schmidt. 2022.
\newblock Exploring the landscape of distributional robustness for question
  answering models.
\newblock \emph{arXiv preprint arXiv:2210.12517}.

\bibitem[{Bentivogli et~al.(2009)Bentivogli, Clark, Dagan, and
  Giampiccolo}]{bentivogli2009fifth}
Luisa Bentivogli, Peter Clark, Ido Dagan, and Danilo Giampiccolo. 2009.
\newblock The fifth pascal recognizing textual entailment challenge.
\newblock In \emph{TAC}.

\bibitem[{Chinea-Rios et~al.(2022)Chinea-Rios, M{\"u}ller, De~la
  Pe{\~n}a~Sarrac{\'e}n, Rangel, and Franco-Salvador}]{chinea2022zero}
Mara Chinea-Rios, Thomas M{\"u}ller, Gretel~Liz De~la Pe{\~n}a~Sarrac{\'e}n,
  Francisco Rangel, and Marc Franco-Salvador. 2022.
\newblock Zero and few-shot learning for author profiling.
\newblock In \emph{International Conference on Applications of Natural Language
  to Information Systems}, pages 333--344. Springer.

\bibitem[{Dagan et~al.(2006)Dagan, Glickman, and Magnini}]{dagan2006pascal}
Ido Dagan, Oren Glickman, and Bernardo Magnini. 2006.
\newblock The pascal recognising textual entailment challenge.
\newblock In \emph{Machine learning challenges workshop}, pages 177--190.
  Springer.

\bibitem[{Devlin et~al.(2018)Devlin, Chang, Lee, and
  Toutanova}]{devlin2018bert}
Jacob Devlin, Ming-Wei Chang, Kenton Lee, and Kristina Toutanova. 2018.
\newblock Bert: Pre-training of deep bidirectional transformers for language
  understanding.
\newblock \emph{arXiv preprint arXiv:1810.04805}.

\bibitem[{Dong et~al.(2021)Dong, Luu, Lin, Yan, and Zhang}]{dong2021should}
Xinshuai Dong, Anh~Tuan Luu, Min Lin, Shuicheng Yan, and Hanwang Zhang. 2021.
\newblock How should pre-trained language models be fine-tuned towards
  adversarial robustness?
\newblock \emph{Advances in Neural Information Processing Systems},
  34:4356--4369.

\bibitem[{Dou et~al.(2019)Dou, Yu, and Anastasopoulos}]{dou2019investigating}
Zi-Yi Dou, Keyi Yu, and Antonios Anastasopoulos. 2019.
\newblock Investigating meta-learning algorithms for low-resource natural
  language understanding tasks.
\newblock \emph{arXiv preprint arXiv:1908.10423}.

\bibitem[{Gao et~al.(2020)Gao, Fisch, and Chen}]{gao2020making}
Tianyu Gao, Adam Fisch, and Danqi Chen. 2020.
\newblock Making pre-trained language models better few-shot learners.
\newblock \emph{arXiv preprint arXiv:2012.15723}.

\bibitem[{Giampiccolo et~al.(2007)Giampiccolo, Magnini, Dagan, and
  Dolan}]{giampiccolo2007third}
Danilo Giampiccolo, Bernardo Magnini, Ido Dagan, and William~B Dolan. 2007.
\newblock The third pascal recognizing textual entailment challenge.
\newblock In \emph{Proceedings of the ACL-PASCAL workshop on textual entailment
  and paraphrasing}, pages 1--9.

\bibitem[{Goldblum et~al.(2020)Goldblum, Fowl, and
  Goldstein}]{goldblum2020adversarially}
Micah Goldblum, Liam Fowl, and Tom Goldstein. 2020.
\newblock Adversarially robust few-shot learning: A meta-learning approach.
\newblock \emph{Advances in Neural Information Processing Systems},
  33:17886--17895.

\bibitem[{Haim et~al.(2006)Haim, Dagan, Dolan, Ferro, Giampiccolo, Magnini, and
  Szpektor}]{haim2006second}
R~Bar Haim, Ido Dagan, Bill Dolan, Lisa Ferro, Danilo Giampiccolo, Bernardo
  Magnini, and Idan Szpektor. 2006.
\newblock The second pascal recognising textual entailment challenge.
\newblock In \emph{Proceedings of the Second PASCAL Challenges Workshop on
  Recognising Textual Entailment}, volume~7.

\bibitem[{Huang et~al.(2017)Huang, Papernot, Goodfellow, Duan, and
  Abbeel}]{huang2017adversarial}
Sandy Huang, Nicolas Papernot, Ian Goodfellow, Yan Duan, and Pieter Abbeel.
  2017.
\newblock Adversarial attacks on neural network policies.
\newblock \emph{arXiv preprint arXiv:1702.02284}.

\bibitem[{Jiang et~al.(2019)Jiang, He, Chen, Liu, Gao, and
  Zhao}]{jiang2019smart}
Haoming Jiang, Pengcheng He, Weizhu Chen, Xiaodong Liu, Jianfeng Gao, and Tuo
  Zhao. 2019.
\newblock Smart: Robust and efficient fine-tuning for pre-trained natural
  language models through principled regularized optimization.
\newblock \emph{arXiv preprint arXiv:1911.03437}.

\bibitem[{Kavumba et~al.(2022)Kavumba, Takahashi, and Oda}]{kavumba2022prompt}
Pride Kavumba, Ryo Takahashi, and Yusuke Oda. 2022.
\newblock Are prompt-based models clueless?
\newblock In \emph{Proceedings of the 60th Annual Meeting of the Association
  for Computational Linguistics (Volume 1: Long Papers)}, pages 2333--2352.

\bibitem[{Lan et~al.(2019)Lan, Chen, Goodman, Gimpel, Sharma, and
  Soricut}]{lan2019albert}
Zhenzhong Lan, Mingda Chen, Sebastian Goodman, Kevin Gimpel, Piyush Sharma, and
  Radu Soricut. 2019.
\newblock Albert: A lite bert for self-supervised learning of language
  representations.
\newblock \emph{arXiv preprint arXiv:1909.11942}.

\bibitem[{Lester et~al.(2021)Lester, Al-Rfou, and Constant}]{lester2021power}
Brian Lester, Rami Al-Rfou, and Noah Constant. 2021.
\newblock The power of scale for parameter-efficient prompt tuning.
\newblock \emph{arXiv preprint arXiv:2104.08691}.

\bibitem[{Li and Liang(2021)}]{li2021prefix}
Xiang~Lisa Li and Percy Liang. 2021.
\newblock Prefix-tuning: Optimizing continuous prompts for generation.
\newblock \emph{arXiv preprint arXiv:2101.00190}.

\bibitem[{Liu et~al.(2022)Liu, Kumar, Liang, and Jia}]{liu2022sample}
Nelson~F Liu, Ananya Kumar, Percy Liang, and Robin Jia. 2022.
\newblock Are sample-efficient nlp models more robust?
\newblock \emph{arXiv preprint arXiv:2210.06456}.

\bibitem[{Liu et~al.(2021)Liu, Zheng, Du, Ding, Qian, Yang, and
  Tang}]{liu2021gpt}
Xiao Liu, Yanan Zheng, Zhengxiao Du, Ming Ding, Yujie Qian, Zhilin Yang, and
  Jie Tang. 2021.
\newblock Gpt understands, too.
\newblock \emph{arXiv preprint arXiv:2103.10385}.

\bibitem[{Liu et~al.(2019)Liu, Ott, Goyal, Du, Joshi, Chen, Levy, Lewis,
  Zettlemoyer, and Stoyanov}]{liu2019roberta}
Yinhan Liu, Myle Ott, Naman Goyal, Jingfei Du, Mandar Joshi, Danqi Chen, Omer
  Levy, Mike Lewis, Luke Zettlemoyer, and Veselin Stoyanov. 2019.
\newblock Roberta: A robustly optimized bert pretraining approach.
\newblock \emph{arXiv preprint arXiv:1907.11692}.

\bibitem[{Madry et~al.(2017)Madry, Makelov, Schmidt, Tsipras, and
  Vladu}]{madry2017towards}
Aleksander Madry, Aleksandar Makelov, Ludwig Schmidt, Dimitris Tsipras, and
  Adrian Vladu. 2017.
\newblock Towards deep learning models resistant to adversarial attacks.
\newblock \emph{arXiv preprint arXiv:1706.06083}.

\bibitem[{Mukherjee and Awadallah(2020)}]{mukherjee2020uncertainty}
Subhabrata Mukherjee and Ahmed Awadallah. 2020.
\newblock Uncertainty-aware self-training for few-shot text classification.
\newblock \emph{Advances in Neural Information Processing Systems},
  33:21199--21212.

\bibitem[{Mukherjee et~al.(2021)Mukherjee, Liu, Zheng, Hosseini, Cheng, Yang,
  Meek, Awadallah, and Gao}]{mukherjee2021clues}
Subhabrata Mukherjee, Xiaodong Liu, Guoqing Zheng, Saghar Hosseini, Hao Cheng,
  Greg Yang, Christopher Meek, Ahmed~Hassan Awadallah, and Jianfeng Gao. 2021.
\newblock Clues: few-shot learning evaluation in natural language
  understanding.
\newblock \emph{arXiv preprint arXiv:2111.02570}.

\bibitem[{Nooralahzadeh et~al.(2020)Nooralahzadeh, Bekoulis, Bjerva, and
  Augenstein}]{nooralahzadeh2020zero}
Farhad Nooralahzadeh, Giannis Bekoulis, Johannes Bjerva, and Isabelle
  Augenstein. 2020.
\newblock Zero-shot cross-lingual transfer with meta learning.
\newblock \emph{arXiv preprint arXiv:2003.02739}.

\bibitem[{Perez et~al.(2021)Perez, Kiela, and Cho}]{perez2021true}
Ethan Perez, Douwe Kiela, and Kyunghyun Cho. 2021.
\newblock True few-shot learning with language models.
\newblock \emph{Advances in Neural Information Processing Systems},
  34:11054--11070.

\bibitem[{Phang et~al.(2018)Phang, F{\'e}vry, and Bowman}]{phang2018sentence}
Jason Phang, Thibault F{\'e}vry, and Samuel~R Bowman. 2018.
\newblock Sentence encoders on stilts: Supplementary training on intermediate
  labeled-data tasks.
\newblock \emph{arXiv preprint arXiv:1811.01088}.

\bibitem[{Rajpurkar et~al.(2016)Rajpurkar, Zhang, Lopyrev, and
  Liang}]{rajpurkar2016squad}
Pranav Rajpurkar, Jian Zhang, Konstantin Lopyrev, and Percy Liang. 2016.
\newblock Squad: 100,000+ questions for machine comprehension of text.
\newblock \emph{arXiv preprint arXiv:1606.05250}.

\bibitem[{Razeghi et~al.(2022)Razeghi, Logan~IV, Gardner, and
  Singh}]{razeghi2022impact}
Yasaman Razeghi, Robert~L Logan~IV, Matt Gardner, and Sameer Singh. 2022.
\newblock Impact of pretraining term frequencies on few-shot reasoning.
\newblock \emph{arXiv preprint arXiv:2202.07206}.

\bibitem[{Schick and Sch{\"u}tze(2020)}]{schick2020s}
Timo Schick and Hinrich Sch{\"u}tze. 2020.
\newblock It's not just size that matters: Small language models are also
  few-shot learners.
\newblock \emph{arXiv preprint arXiv:2009.07118}.

\bibitem[{Schick and Sch{\"u}tze(2021)}]{schick2021exploiting}
Timo Schick and Hinrich Sch{\"u}tze. 2021.
\newblock Exploiting cloze-questions for few-shot text classification and
  natural language inference.
\newblock In \emph{Proceedings of the 16th Conference of the European Chapter
  of the Association for Computational Linguistics: Main Volume}, pages
  255--269.

\bibitem[{Schick and Sch{\"u}tze(2022)}]{schick2022true}
Timo Schick and Hinrich Sch{\"u}tze. 2022.
\newblock True few-shot learning with prompts—a real-world perspective.
\newblock \emph{Transactions of the Association for Computational Linguistics},
  10:716--731.

\bibitem[{Socher et~al.(2013)Socher, Perelygin, Wu, Chuang, Manning, Ng, and
  Potts}]{socher2013recursive}
Richard Socher, Alex Perelygin, Jean Wu, Jason Chuang, Christopher~D Manning,
  Andrew~Y Ng, and Christopher Potts. 2013.
\newblock Recursive deep models for semantic compositionality over a sentiment
  treebank.
\newblock In \emph{Proceedings of the 2013 conference on empirical methods in
  natural language processing}, pages 1631--1642.

\bibitem[{Tam et~al.(2021)Tam, Menon, Bansal, Srivastava, and
  Raffel}]{tam2021improving}
Derek Tam, Rakesh~R Menon, Mohit Bansal, Shashank Srivastava, and Colin Raffel.
  2021.
\newblock Improving and simplifying pattern exploiting training.
\newblock \emph{arXiv preprint arXiv:2103.11955}.

\bibitem[{Wang et~al.(2018)Wang, Singh, Michael, Hill, Levy, and
  Bowman}]{wang2018glue}
Alex Wang, Amanpreet Singh, Julian Michael, Felix Hill, Omer Levy, and Samuel~R
  Bowman. 2018.
\newblock Glue: A multi-task benchmark and analysis platform for natural
  language understanding.
\newblock In \emph{International Conference on Learning Representations}.

\bibitem[{Wang et~al.(2021{\natexlab{a}})Wang, Xu, Wang, Gan, Cheng, Gao,
  Awadallah, and Li}]{wang2021adversarial}
Boxin Wang, Chejian Xu, Shuohang Wang, Zhe Gan, Yu~Cheng, Jianfeng Gao,
  Ahmed~Hassan Awadallah, and Bo~Li. 2021{\natexlab{a}}.
\newblock Adversarial glue: A multi-task benchmark for robustness evaluation of
  language models.
\newblock In \emph{Thirty-fifth Conference on Neural Information Processing
  Systems Datasets and Benchmarks Track (Round 2)}.

\bibitem[{Wang et~al.(2021{\natexlab{b}})Wang, Fang, Khabsa, Mao, and
  Ma}]{wang2021entailment}
Sinong Wang, Han Fang, Madian Khabsa, Hanzi Mao, and Hao Ma.
  2021{\natexlab{b}}.
\newblock Entailment as few-shot learner.
\newblock \emph{arXiv preprint arXiv:2104.14690}.

\bibitem[{Wang et~al.(2021{\natexlab{c}})Wang, Mukherjee, Liu, Gao, Awadallah,
  and Gao}]{wang2021list}
Yaqing Wang, Subhabrata Mukherjee, Xiaodong Liu, Jing Gao, Ahmed~Hassan
  Awadallah, and Jianfeng Gao. 2021{\natexlab{c}}.
\newblock List: Lite self-training makes efficient few-shot learners.
\newblock \emph{arXiv preprint arXiv:2110.06274}.

\bibitem[{Wang and Chang(2022)}]{wang2022toxicity}
Yau-Shian Wang and Yingshan Chang. 2022.
\newblock Toxicity detection with generative prompt-based inference.
\newblock \emph{arXiv preprint arXiv:2205.12390}.

\bibitem[{Williams et~al.(2017)Williams, Nangia, and
  Bowman}]{williams2017broad}
Adina Williams, Nikita Nangia, and Samuel~R Bowman. 2017.
\newblock A broad-coverage challenge corpus for sentence understanding through
  inference.
\newblock \emph{arXiv preprint arXiv:1704.05426}.

\bibitem[{Wortsman et~al.(2022)Wortsman, Ilharco, Kim, Li, Kornblith, Roelofs,
  Lopes, Hajishirzi, Farhadi, Namkoong et~al.}]{wortsman2022robust}
Mitchell Wortsman, Gabriel Ilharco, Jong~Wook Kim, Mike Li, Simon Kornblith,
  Rebecca Roelofs, Raphael~Gontijo Lopes, Hannaneh Hajishirzi, Ali Farhadi,
  Hongseok Namkoong, et~al. 2022.
\newblock Robust fine-tuning of zero-shot models.
\newblock In \emph{Proceedings of the IEEE/CVF Conference on Computer Vision
  and Pattern Recognition}, pages 7959--7971.

\bibitem[{Xie et~al.(2020)Xie, Dai, Hovy, Luong, and Le}]{xie2020unsupervised}
Qizhe Xie, Zihang Dai, Eduard Hovy, Thang Luong, and Quoc Le. 2020.
\newblock Unsupervised data augmentation for consistency training.
\newblock \emph{Advances in Neural Information Processing Systems},
  33:6256--6268.

\bibitem[{Yin et~al.(2020)Yin, Rajani, Radev, Socher, and
  Xiong}]{yin2020universal}
Wenpeng Yin, Nazneen~Fatema Rajani, Dragomir Radev, Richard Socher, and Caiming
  Xiong. 2020.
\newblock Universal natural language processing with limited annotations: Try
  few-shot textual entailment as a start.
\newblock \emph{arXiv preprint arXiv:2010.02584}.

\bibitem[{Zhao et~al.(2021)Zhao, Wallace, Feng, Klein, and
  Singh}]{zhao2021calibrate}
Zihao Zhao, Eric Wallace, Shi Feng, Dan Klein, and Sameer Singh. 2021.
\newblock Calibrate before use: Improving few-shot performance of language
  models.
\newblock In \emph{International Conference on Machine Learning}, pages
  12697--12706. PMLR.

\bibitem[{Zheng et~al.(2021)Zheng, Zhou, Qian, Ding, Liao, Li, Salakhutdinov,
  Tang, Ruder, and Yang}]{zheng2021fewnlu}
Yanan Zheng, Jing Zhou, Yujie Qian, Ming Ding, Chonghua Liao, Jian Li, Ruslan
  Salakhutdinov, Jie Tang, Sebastian Ruder, and Zhilin Yang. 2021.
\newblock Fewnlu: Benchmarking state-of-the-art methods for few-shot natural
  language understanding.
\newblock \emph{arXiv preprint arXiv:2109.12742}.

\end{thebibliography}
\bibliographystyle{acl_natbib}

\end{document}